\documentclass[sigconf, twocolumn]{acmart}

\AtBeginDocument{%
  \providecommand\BibTeX{{%
    \normalfont B\kern-0.5em{\scshape i\kern-0.25em b}\kern-0.8em\TeX}}}

\setcopyright{acmlicensed}
\copyrightyear{2024}
\acmYear{2024}
\acmDOI{XXXXXXX.XXXXXXX}

\acmConference[ACM SIGCAS/SIGCHI]{Make sure to enter the correct
  conference title from your rights confirmation emai}{July 8-11, 2024}{IIIT-Delhi, India}

\acmISBN{978-1-4503-XXXX-X/18/06}

\usepackage{subcaption}
\usepackage{ragged2e}
\usepackage{caption}

\begin{document}

\title{HumekaFL: Automated Detection of Neonatal Asphyxia Using Federated Learning}


\author{Pamely Zantou}
\email{pzantou@andrew.cmu.edu}
\affiliation{
  \institution{Carnegie Mellon University}
  \city{Kigali}
  \country{Rwanda}
}

\author{Blessed Guda}
\email{blessedg@andrew.cmu.edu}
\affiliation{
  \institution{Carnegie Mellon University}
  \city{Kigali}
  \country{Rwanda}
}

\author{Bereket Retta}
\email{bretta@andrew.cmu.edu}
\affiliation{
  \institution{Carnegie Mellon University}
  \city{Kigali}
  \country{Rwanda}
}

\author{Gladys Inabeza}
\email{ginabeza@alumni.cmu.edu}
\affiliation{
  \institution{Carnegie Mellon University}
  \city{Kigali}
  \country{Rwanda}
}

\author{Carlee Joe-Wong}
\email{cjoewong@andrew.cmu.edu}
\affiliation{
  \institution{Carnegie Mellon University}
  \city{Pittsburgh}
  \country{USA}
}

\author{Assane Gueye}
\email{assaneg@andrew.cmu.edu}
\affiliation{
  \institution{Carnegie Mellon University}
  \city{Kigali}
  \country{Rwanda}
}


\begin{abstract}
{
    Birth Apshyxia (BA) is a severe condition characterized by insufficient supply of oxygen to a newborn during the delivery. BA is one of the primary causes of neonatal death in the world. Although there has been a decline in neonatal deaths over the past two decades, the developing world, particularly sub-Saharan Africa, continues to experience the highest under-five (<5) mortality rates. While evidence-based methods are commonly used to detect BA in African healthcare settings, they can be subject to physician errors or delays in diagnosis, preventing timely interventions. Centralized Machine Learning (ML) methods demonstrated good performance in early detection of BA but require sensitive health data to leave their premises before training, which does not guarantee privacy and security. Healthcare institutions are therefore reluctant to adopt such solutions in Africa. To address this challenge, we suggest a federated learning (FL)-based software architecture, a distributed learning method that prioritizes privacy and security by design. We have developed a user-friendly and cost-effective mobile application embedding the FL pipeline for early detection of BA. Our Federated SVM model outperformed centralized SVM pipelines and Neural Networks (NN)-based methods in the existing literature. 

}
\end{abstract}



\keywords{Neonatal Mortality, Birth Asphyxia, Machine Learning, Federated Learning, Mobile Application}


\maketitle

\section{ML-based Early Detection of BA Faces Deployment Challenges}
{
    Early detection of BA and timely intervention can facilitate full recovery for infants with mild or moderate asphyxia. Conversely, delayed detection results in prolonged oxygen deprivation, leading to permanent injuries that may impact various organs such as the brain, heart, lungs, kidneys, and bowels or other organs. 

    Research in the healthcare community has also established that a baby's cry immediately after birth plays a crucial role in language acquisition \cite{kheddache2019identification}. An infant crying vigorously after birth accounts for an APGAR (Appearance, Pulse, Grimace, Activity, and Respiration) score of 2 \cite{kheddache2019identification}. Consequently, a weak or delayed cry can serve as an indicator of birth asphyxia. Moreover, babies' cries have been used extensively to build technological solutions to detect BA. Ubenwa, developed by \cite{onu2017ubenwa} stands as a leading ML-powered diagnostic mobile application leveraging newborn cries for BA detection. 
 
    Research addressing automated diagnosis of neonatal asphyxia has explored a range of methods and modalities. Automatic speech recognition (ASR) stands out as a widely used and successful approach. Various machine learning (ML) methods have been applied in ASR research for BA  classification. Traditional ML classifiers like Support Vector Machine (SVM), K-Nearest Neighbor (KNN) and deep learning (DL) methods such as Convolutional Neural Network (CNN) or Graph Neural Networks (GNN) have demonstrated good performance on BA detection \cite{ji2021review}.   
    
    While many research efforts focus on cry analysis and classification, only a few have been deployed in healthcare facilities to aid healthcare professionals in improving BA diagnosis, particularly in the developing world. For example, Ubenwa has been tested in hospitals in Nigeria and Canada but has yet to achieve widespread adoption and deployment across sub-Saharan Africa's healthcare space. Three deployment challenges might impede the integration of ML-based solutions like Ubenwa into the sub-Saharan healthcare facilities:    
    \begin{enumerate}
        \item Privacy and security concerns due to the high sensitivity of healthcare data which centralized methods of leveraging ML pipelines do not always ensure primarily because health data have to leave their premises. HumekaFL addresses this by leveraging a federated learning pipeline.  
        \item The lack of computing resources to train large ML models. HumekaFL solves this by using smaller models and harnessing the power of mobile devices. 
        \item The absence of user-friendly ML-based solutions which do not require prior knowledge to assist health professionals and verified parents or caregivers in identifying BA. 
    \end{enumerate}

    To be addressed effectively, these problems require health professionals, policymakers and technologists to collaborate and provide sustainable and responsible solutions. We propose HumekaFL \footnote{This research is named HumekaFL, a blend of ``Humeka'', which means ``Breathe'' in Kinyarwanda and "FL", an abbreviation for Federated Learning. Kinyarwanda is the national language spoken in Rwanda and in some neighboring countries in the East African region. Therefore, HumekaFL represents a federated learning solution geared toward \textbf{helping infants breathe}.}, an affordable and user-friendly (requiring no prior training) software application that utilizes machine learning to detect BA. In doing so, we design an end-to-end system architecture that can be readily deployed by healthcare facilities and respects the privacy needs of their sensitive data. To meet these challenges, we use a federated learning (FL) architecture that does not require sensitive data to leave its premises. We thus address the three AI deployment challenges raised above: by running on commodity hardware and requiring no prior training, HumekaFL does not require specialized equipment or advanced skills.

}

\section{Early Detection of BA Using Federated Learning}
{
    \par{ \textbf{System Architecture}. HumekaFL leverages an ML pipeline that employs FL to detect BA. Federated learning (FL) is a distributed learning  approach which enables training models across several clients devices using their local data while ensuring data privacy (Figure \ref{diagnostic-process})\cite{mcmahan2017communication}. 

    HumekaFL uses a cross-silo and centralized federated learning architecture to train a model and make inference on local data. In particular, multiple local clients \--- they can be mobile devices, computers or more powerful computing devices like servers, available in health facilities \--- in a simulated environment, each collect data and train local models based on this data. The models are periodically aggregated at a central server, which then pushes the aggregated model or global model back to the clients for subsequent updates. Typically, stochastic gradient descent or one of its variants is used for the model updates. The global model is obtained using an aggregation algorithm such as Federated Averaging (FedAvg) algorithms \cite{mcmahan2017communication}, FedProx \cite{li2020federated} or any other model aggregation technique. 
        
    \begin{figure}[!hbpt]
        \centering
        \includegraphics[scale=0.2]{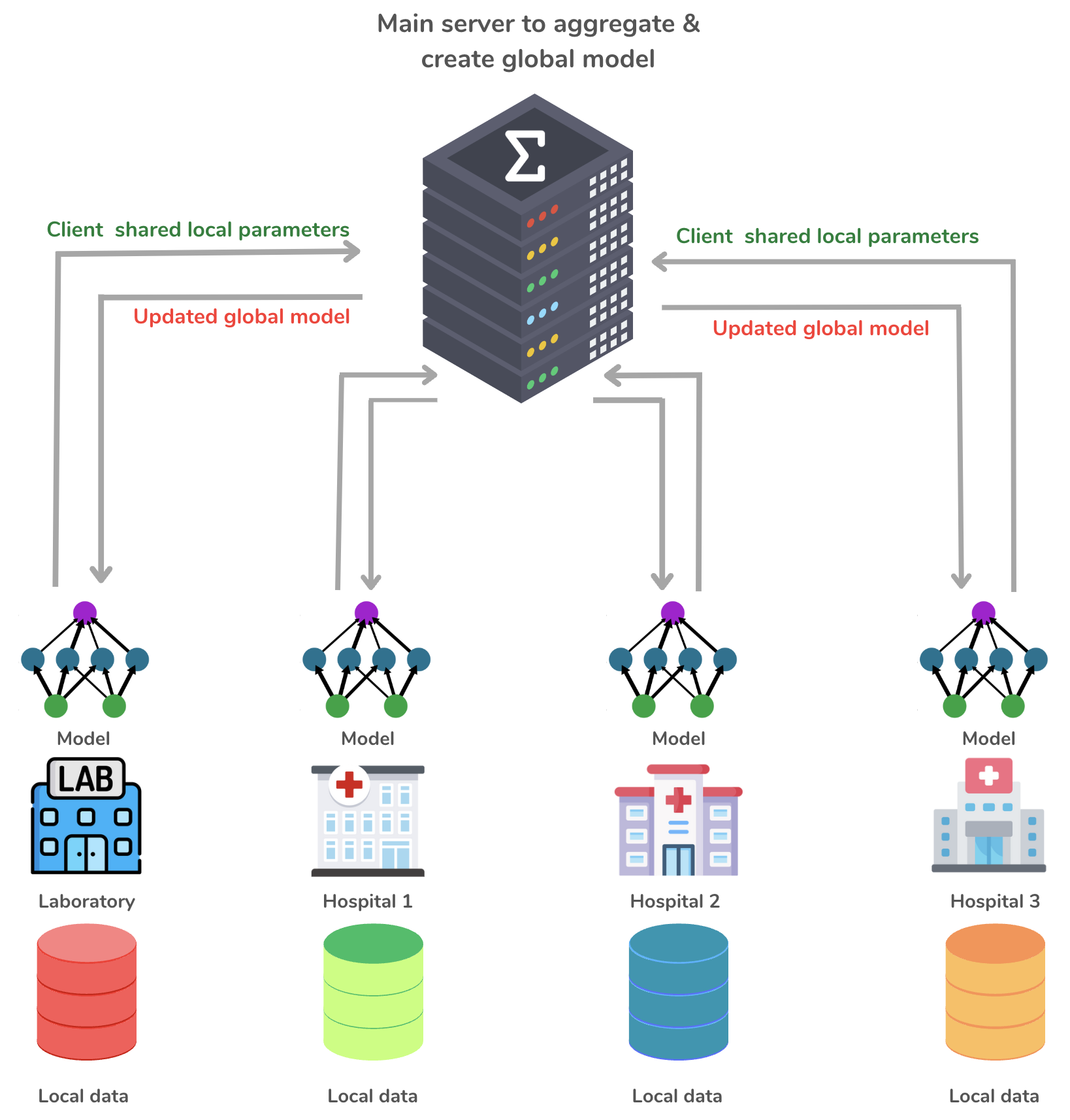}
        \caption{System architecture}
        \label{sys-architecture}
    \end{figure}
    }
    
    \par{\label{dataprocessing} \noindent \textbf{Data Pre-processing}. HumekaFL uses the Baby Chillanto Dataset courtesy of the National Institute of Astrophysics and Optical Electronics, CONACYT, Mexico\cite{reyes2008evolutionary} to train its ML pipeline. It is a small dataset which comprises 1,049 recordings from healthy neonates, 870 cry recordings from deaf infants, and 340 cry recordings from infants affected by perinatal asphyxia. The database encompasses five distinct types of cry signals, including those from deaf infants, infants with asphyxia, normal cries, and cries indicative of pain. Each audio recording is one second (1s) in duration and sampled at frequencies ranging from 8kHz to 16kHz, utilizing 16-bit PCM encoding. We use Mel-Frequency Cepstral Coefficients (MFCCs) to extract features from the newborns' cries. Forty (40) MFCCs were used to represent every 1000 ms cry clip from the Baby Chillanto database. 

    The Baby Chillanto Infant Cry database is small in size. This can present a challenge for ML/DL pipelines that typically benefit from larger datasets to enhance learning speed and performance. Consequently, we augmented the data. In our data augmentation process, we chose to bootstrap the dataset using two techniques: \textit{tanh()} distortion and room reverberation. As a result of this augmentation, our dataset now comprises 1521 normal data points and 1028 asphyxiated data points.

    \textit{Tanh() distortion} adds a rounded soft clipping to the audio signal. The distortion amount is proportional to the loudness of the input and its gain. The \textit{tanh} function is symmetric. After adding \textit{a} distortion, the positive and negative parts of the signal are squashed. \\
    \indent Tanh function without distortion:
    \begin{equation}
        \frac{sinh(x)}{cosh(x)} = \frac{e^x - e^{-x}}{e^x + e^{-x}}    
    \end{equation}

   Tanh function with distortion G added: 
    \begin{equation}
        \frac{e^{x*(a + G)} - e^{x*(b - G)}} {e^{x*G} - e^{x*-G}}
    \end{equation}

    \textit{Room reverberation} consists in creating audio signals by simulating the reverberation of sound inside rooms. To perform room reverberation, we use Room Impulse Response (RIR). The RIR is cleaned up and used to extract the main impulse, normalize the signal power, then flip along the time axis. Finally, the signal is convolved with a RIR filter.

    The last step in pre-processing the training data is extracting the most relevant MFCCs through a Random Forest Classifier.
       
    While our model is fed with the MFCCs extracted from the audio clips during training, we also pre-process the audio recorded with our mobile application before performing inference. Our mobile application records 10s of babies cry and passes it to our model for inference. The recorded audio signal might be noisy since it can be recorded either from the hospital by health professionals or anywhere else noisy by mothers or caregivers. Before sending the newborn cry to the pipeline we first use Google WebRTC Voice Activity Detector (VAD) \cite{webrtc} algorithm to capture part of the signal with human speech (the baby's cry in our case). Afterward, a Butterworth band-pass filter is applied to the signal to remove noise and extract a clean signal ready to be passed to the \textit{on-device} model for classification. This clean data can then be utilized to re-train and refine the model. Although the current pipeline doesn't incorporate the collected data for model updates, in future work we plan to integrate this data for continuous improvement of the model.
    

    \begin{figure}[!hbpt]
        \centering
        \includegraphics[scale=0.2175]{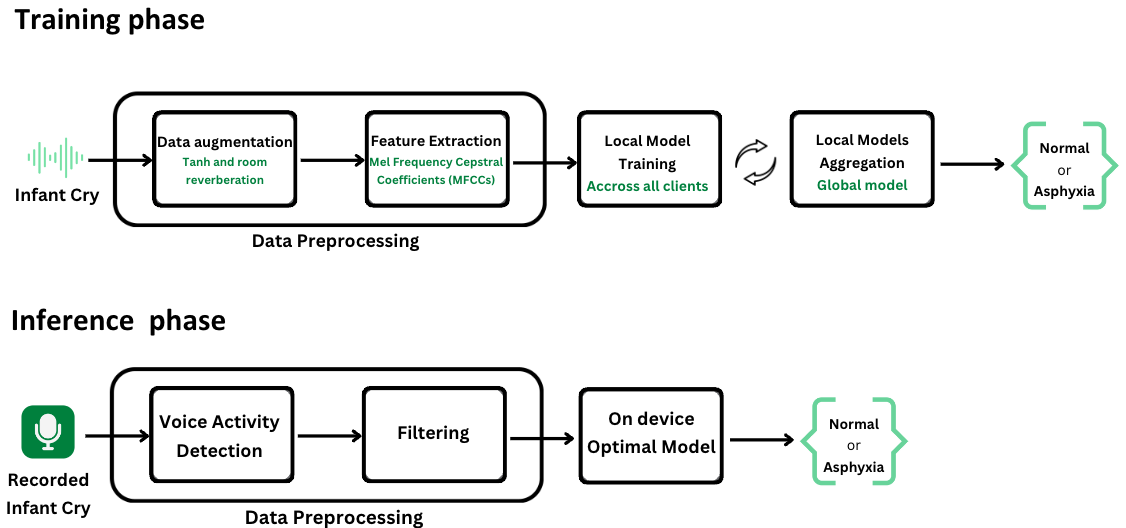}
        \caption{HumekaFL Diagnostic Process}
        \label{diagnostic-process}
    \end{figure}

    }

    \par{ \noindent \textbf{Federated SVM (FedSVM)}. To classify newborns' cries in a federated setting, we first decide the ML model to employ in training our data on distributed clients. HumekaFL leverages Support Vector Machine (SVM) to train the Baby Chillanto Dataset on ten (10) virtual silos (hospitals). There are few important considerations justifying this choice: (1) the size of the training dataset, which is relatively small, (2) computation efficiency in a resource-constrained environment (number of trainable parameters, speed/time and memory), (3) compliance to Stochastic Gradient Descent (SGD) optimization algorithm and (4) overall performance on the cry classification task based on the existing literature \cite{onu2017ubenwa}. SVM is known for its ability to efficiently learn complex decision boundaries from small high-dimensional dataset such as the Baby Chillanto. Furthermore, SVM is also recognized to use a very small number of parameters and amount of memory to learn decision boundaries, unlike state-of-the-art neural networks requiring heavy computation resources to learn the right signal in making decision. SVM \--- unlike Decision Trees \--- can use SGD to optimize its objective function. This is paramount for the federated aggregation of clients' model gradients on the orchestration server in FL. Finally, SVM has demonstrated good performance on newborns' cry classification \cite{ji2021review, onu2017ubenwa}. 
    
    HumekaFL's FedSVM consists in several client-server communication rounds in the training phase. Each round consists in local trainings of a shared SVM model followed by a transmission of selected silo's gradients to the server. Client-server communication is coordinated by the FedAvg algorithm. 

    \textbf{Local training of SVM model}. HumekaFL trains on each silo a feature-label (cry - asphyxia status) pairs $(x^{k}, y^{k}) \epsilon R^{D}$ x $\{-1, 1\}$ where k = 1, ....., N, a linear kernel SVM classifier $h(x)= w^Tx$ which finds the regularized empirical risk or the averaged loss function $l(h(x), y)$  over the entire dataset (\ref{empirical_risk}) and minimize it (\ref{optmization_problem}). 
    \begin{equation}
        F(w) = \frac{\lambda}{2} \|\mathbf{w}\|^2_2 + \frac{1}{n}\sum^n_{i=1}l(w^Tx_{i},y_{i}) \label{empirical_risk}
    \end{equation}
    \begin{equation}
        \widehat{w} = \underset{w}{\operatorname{argmin}} F(w) \label{optmization_problem}
    \end{equation}
    where $\lambda$ is a parameter controlling the strength of the regularization and $l(w^Tx_{i},y_{i}) = max(0, 1 - w^{T}xy)$ which is the Hinge loss function. To solve the optimization problem in (\ref{optmization_problem}), we use the Adam optimization algorithm \cite{kingma2014adam}. 
    }
    
    \textbf{Global SVM model}. During each client-server communication round in the training phase, FedAvg randomly selects a group of clients from the available pool $S_{t}$ of clients. To create the global SVM model, the parameters of each chosen client are weighted and averaged, with the weighting factor being proportional to the volume of data each client possesses. This translates into $w_{glob}^{t+1} = \sum_{k \in S_{t}} \frac{n_k}{n} w_{k}^{t+1}$ where $k \in S_{t}$ represents the $k^{th}$ silo of the selected silos; $\frac{n_k}{n}$ the ratio of data quantity on client k to the total data quantity; $w_{k}^t+1$ denotes the updated model of client $k$ and $w_{glob}^{t+1}$ the updated aggregated model. Client-server communication rounds continue until the global SVM model converges. 

    \textbf{Inference}. Upon training, our FedSVM is integrated to the mobile application to perform inference. Figure \ref{diagnostic-process} shows the inference pipeline. We record the baby's voice using the mobile application. The recorded signal is then pre-processed \--- as described in the \textit{Data Pre-processing} section \ref{dataprocessing} \--- followed by the diagnostic result thanks to the \textit{on-device} FedSVM model. 
}

\section{Experiments and Results}
{
    We conducted experiments to validate our FedSVM pipeline's performance and embed our model on a hybrid mobile application. Figure \ref{fig:mobile-app} shows HumekaFL mobile application's screens. Our approach consists in first training a SVM pipeline on the Baby Chillanto dataset. We compare our results with the SVM pipeline developed for Ubenwa, the automated diagnosis tool that uses the same training dataset \cite{onu2017ubenwa}. Our SVM model outperformed Ubenwa’s SVM model, achieving a sensitivity of \textbf{0.91}, a specificity of \textbf{0.96} and an unweighted average recall (UAR) of \textbf{0.93}. In contrast, the Ubenwa SVM model exhibited a sensitivity of 0.85, a specificity of 0.89 and a UAR of 0.87. The main difference between our SVM model and the Ubenwa model is the data augmentation, hyperparameter tuning, feature selection of MFCCs and the federated training setup. 

    \begin{table}[]
        \centering
        \begin{tabular}{llr}
            \toprule
            \textbf{Models} & \textbf{Metric}  & \textbf{Metric Values}\\ \midrule
            Ubenwa SVM & Sensitivity &  0.85      \\
            Ubenwa SVM & Specificity &  0.89  \\ 
            Ubenwa SVM & UAR &  0.865  \\ 
            HumekaFL SVM & Sensitivity &  \textbf{0.91}  \\ 
            HumekaFL SVM & Specificity &  \textbf{ 0.96}  \\ 
            HumekaFL SVM & UAR &  \textbf{0.93}  \\
            \bottomrule   
        \end{tabular}
            \caption{HumekaFL SVM model performance}
            \label{tab:svm-performance}
    \end{table}

    \begin{figure}[!hbpt]
          \begin{subfigure}{0.15\textwidth}
            \centering
            \includegraphics[width=\linewidth]{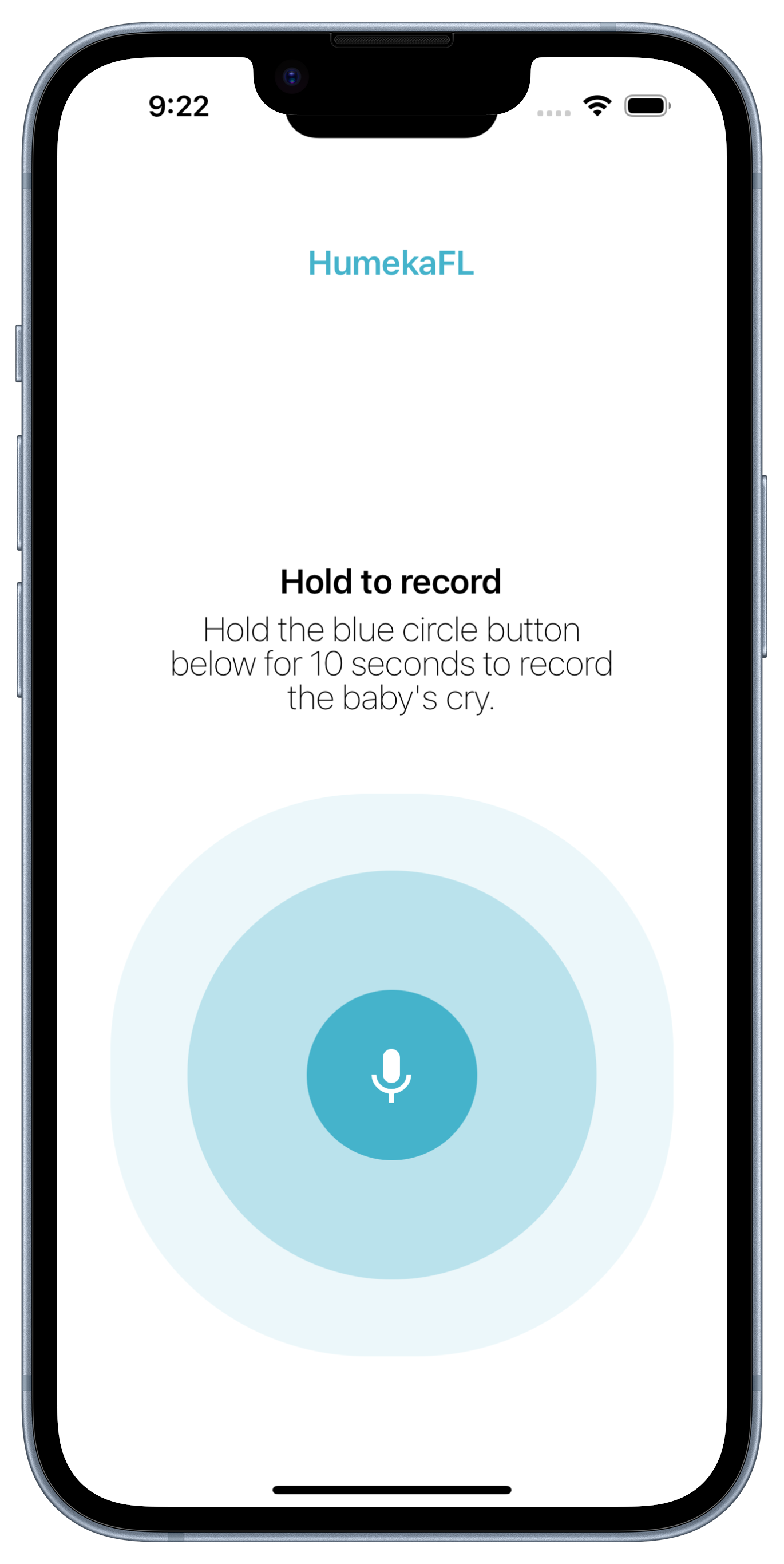}
            \caption{Recording}
            \label{subfig:recording}
          \end{subfigure}
            \begin{subfigure}{0.15\textwidth} 
            \centering
            \includegraphics[width=\linewidth]{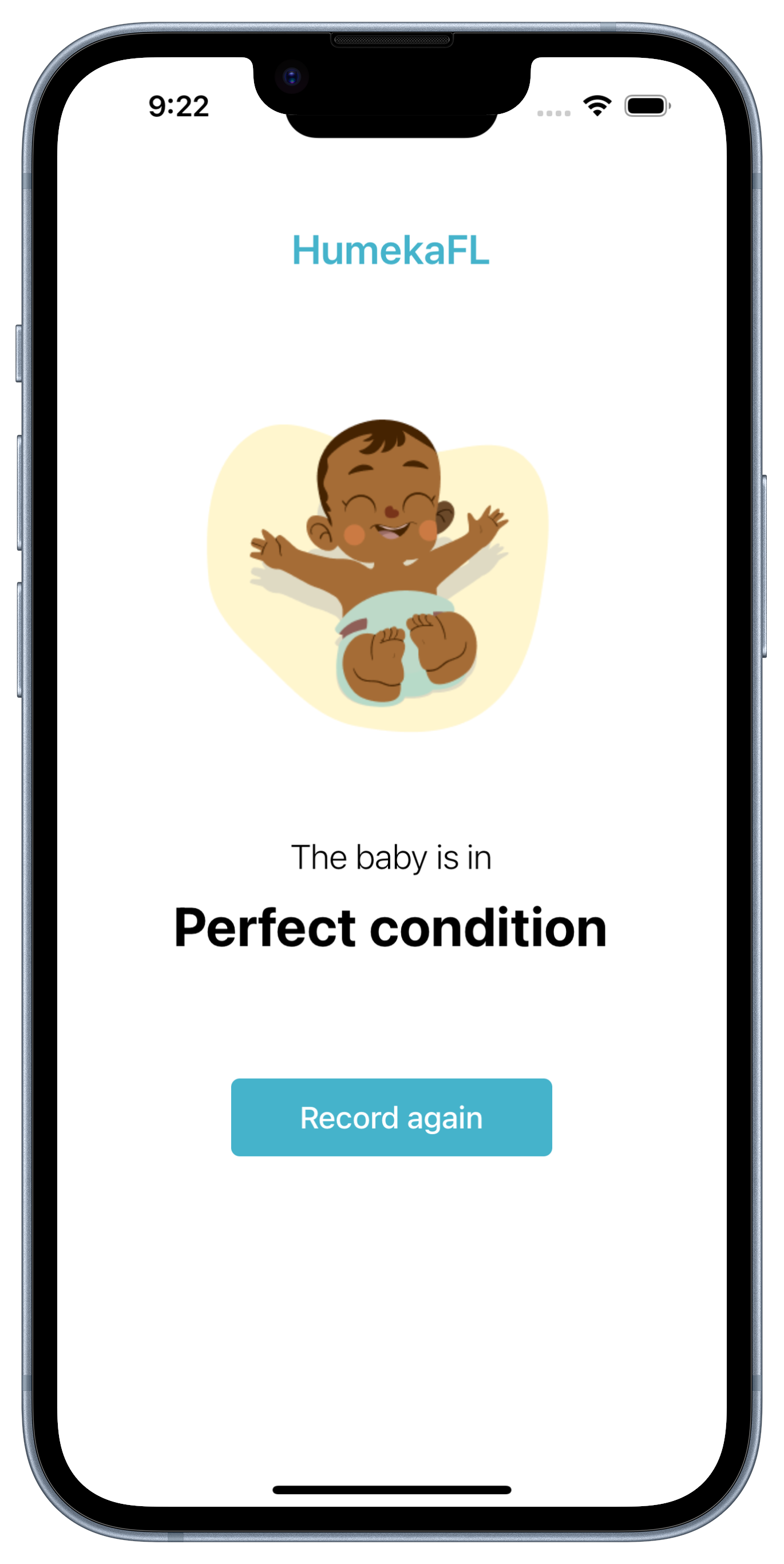}
            \caption{Perfect baby}
            \label{subfig:perfect}
          \end{subfigure}%
          \begin{subfigure}{0.15\textwidth}
            \centering
            \includegraphics[width=\linewidth]{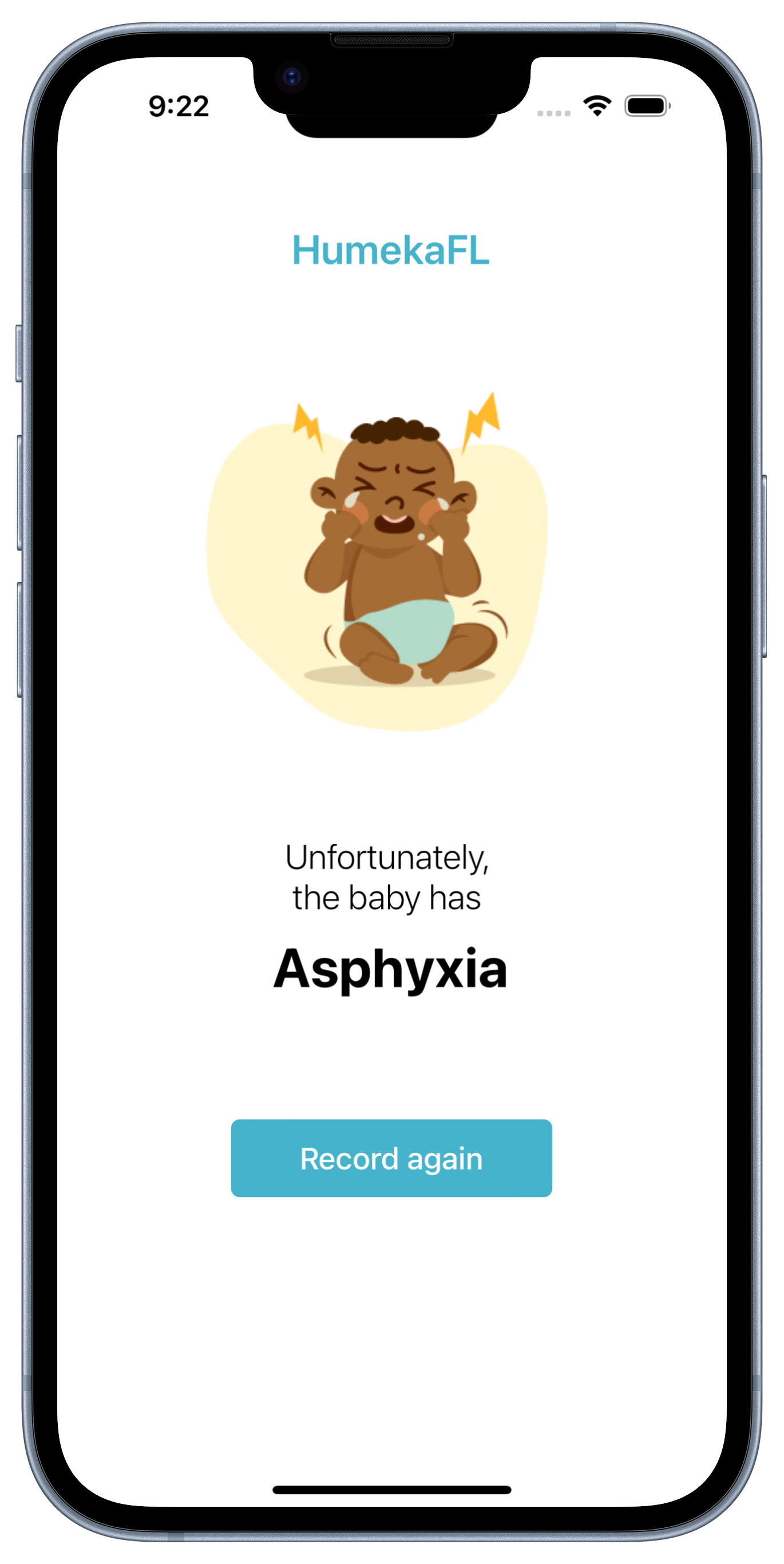}
            \caption{Asphyxiated baby}
            \label{subfig:asphyxia}
        \end{subfigure}
        \caption{HumekaFL mobile application's screens}
        \label{fig:mobile-app}
    \end{figure}

    Our second set of experiments consist in training ten (10) clients over fifty (50) rounds. Each client trains its local model for five (5) epochs. HumekaFL FedSVM model demonstrated a good performance of \textbf{95.88\% }average accuracy following training. However, we need to conduct more experiments with physical healthcare clients to validate these preliminary results. Moreover, further experiments need to be using African health data to prevent biases and enhance our model performance in African healthcare settings. 

    \begin{figure}[!hbpt]
      \begin{subfigure}[b]{0.45\columnwidth}
        \includegraphics[width=\linewidth]{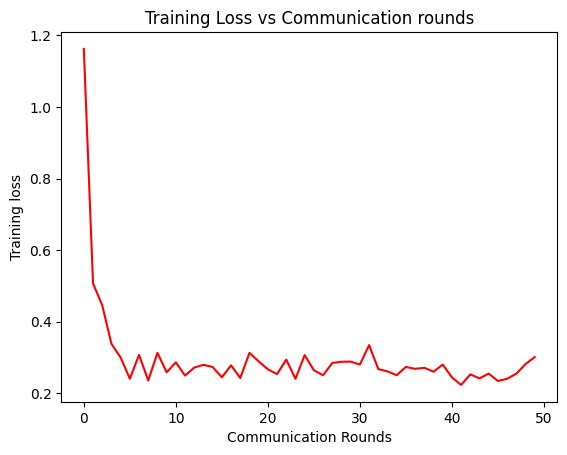}
        \caption{Training Loss vs Communication rounds}
        \label{fig:train-loss-vs-com-rounds}
      \end{subfigure}
      \hfill 
      \begin{subfigure}[b]{0.45\columnwidth}
        \includegraphics[width=\linewidth]{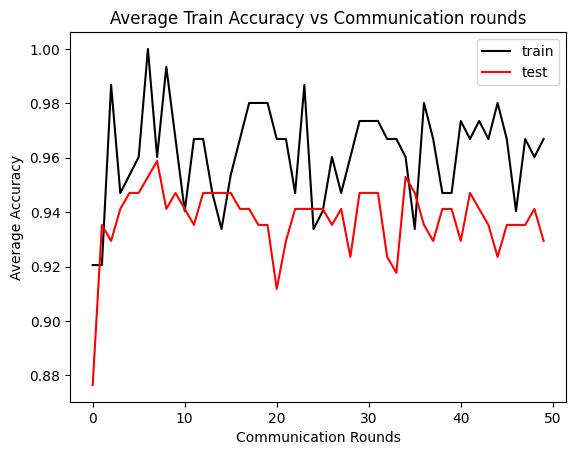}
        \caption{Average Train Accuracy vs Communication rounds}
        \label{fig:avg-vs-com-rounds}
      \end{subfigure}
    \end{figure}


}

\section{Conclusion}
{
    In this work, we propose HumekaFL an FL-based mobile application for early detection of newborn birth asphyxia. Our SVM-based federated pipeline, FedSVM, outperformed existing centralized SVM and NN-based models using the Baby Chillanto dataset. Data augmentation, hyperparameter tuning and feature selection in our federated environment revealed to be successful, but in future work, we will refine our current prototype by collaborating with Rwandan hospitals for African data collection. This will help us conduct clinical trials in health facilities and avoid possible biases resulting from not using African data for training our FedSVM pipeline. Additionally, we will incorporate privacy concerns through differential privacy suggested by \cite{dwork2006differential}. Furthermore, we will enhance the functionality of our FedSVM pipeline to securely utilize data from our mobile application for retraining purposes.
}

\section{Acknowledgements}

    The Baby Chillanto Data Base is a property of the Instituto Nacional de Astrofisica Optica y Electronica – CONACYT, Mexico. We like to thank Dr. Carlos A. Reyes-Garcia, Dr. Emilio Arch-Tirado and his INR-Mexico group, and Dr. Edgar M. Garcia-Tamayo for their dedication of the collection of the Infant Cry database.
\bibliographystyle{ACM-Reference-Format}
\bibliography{humekafl}

\end{document}